\documentclass[runningheads]{llncs}
\usepackage[T1]{fontenc}
\usepackage{arydshln}
\usepackage{hyperref}
\usepackage{graphicx}
\usepackage[status=draft,nomargin,inline,lang=spanish]{fixme}
\usepackage{xcolor}
\newcommand{\eg}[0]{\textit{e.g., }}

\begin{document}
%
\title{Pareto-Optimized Open-Source LLMs for Healthcare via Context Retrieval}
\titlerunning{Cheap, Fast and Open LLMs via Context Retrieval}
%
\author{Jordi Bayarri-Planas\inst{1}\orcidID{0009-0005-1968-3467} \and
Ashwin Kumar Gururajan\inst{1}\orcidID{0000-0002-9246-4552} \and
Dario Garcia-Gasulla\inst{1}\orcidID{0000-0001-6732-5641}}
\authorrunning{Jordi Bayarri et al.}
%
\institute{Barcelona Supercomputing Center (BSC)}
\maketitle              
\begin{abstract}
This study leverages optimized context retrieval to enhance open-source Large Language Models (LLMs) for cost-effective, high performance healthcare AI. We demonstrate that this approach achieves state-of-the-art accuracy on medical question answering at a fraction of the cost of proprietary models, significantly improving the cost-accuracy Pareto frontier on the MedQA benchmark. Key contributions include: (1) OpenMedQA, a novel benchmark revealing a performance gap in open-ended medical QA compared to multiple-choice formats; (2) a practical, reproducible pipeline for context retrieval optimization; and (3) open-source resources (\href{https://github.com/HPAI-BSC/prompt_engine}{\texttt{Prompt Engine}}, CoT/ToT/Thinking databases) to empower healthcare AI development. By advancing retrieval techniques and QA evaluation, we enable more affordable and reliable LLM solutions for healthcare. All the materials have been made public \href{https://huggingface.co/collections/HPAI-BSC/medical-context-retrieval-rag-67b0e0b0589983db691217cd}{here}.
\keywords{Open-Source LLMs \and Healthcare AI \and Context Retrieval \and OpenMedQA \and Pareto Frontier \and Cost-Efficiency \and Medical Question Answering}
\end{abstract}

\begin{figure}
    \centering
    \includegraphics[width=\linewidth]{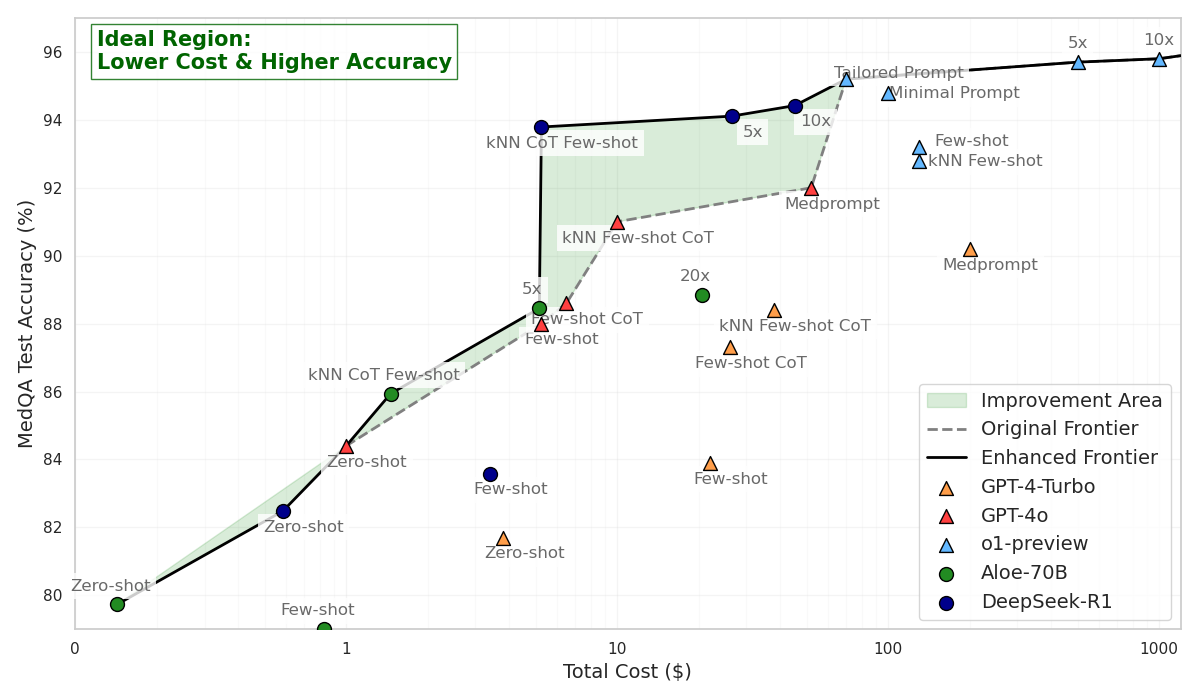}
    \caption{\textbf{Enhanced Pareto Frontier of Accuracy vs. Cost on MedQA}. The solid line represents the improved efficiency frontier, demonstrably surpassing the original Pareto frontier (dashed line). Circular markers indicate open-source models, while triangles represent closed models. The green shaded area visually highlights the region of significant cost-effective accuracy gains.}
    \label{fig:pareto}
\end{figure}

\section{Introduction}

Large Language Models (LLMs) are poised to revolutionize healthcare, offering unprecedented capabilities for natural language understanding and generation in medical applications. Realizing this potential in practice requires reaching high levels of performance and reliability, as demanded by such a sensitive domain. While proprietary LLMs often dominate benchmarks, their high operational costs and limited accessibility restrict widespread adoption, particularly in resource-constrained healthcare settings. This cost barrier is further amplified when considering the need for robust factuality and minimal hallucination in medical contexts, often necessitating more expensive models and complex inference strategies.

This study directly addresses the challenge of cost-effective, high-performance healthcare AI by demonstrating the transformative power of optimized context retrieval in conjunction with open-source LLMs. Our central thesis is that by strategically enhancing open models with optimized retrieval mechanisms, we can achieve performance levels comparable to costly proprietary solutions but at a fraction of the computational expense. Building upon the framework of Pareto frontiers for evaluating run-time strategies, we empirically validate a significant improvement in the cost-accuracy trade-off for medical question answering. As illustrated in Figure~\ref{fig:pareto}, our approach extends the Pareto frontier on the MedQA benchmark, enabling open-source models to operate in a new efficiency regime that balances high accuracy with reduced computational costs.

Our work offers a practical pathway toward reducing barriers to high-quality healthcare AI, shifting the focus from solely relying on ever-larger, more expensive models to leveraging efficient, cost-optimized architectures and techniques. Contributions of this work are:

\begin{itemize}
    \item \textbf{Practical guide for cost-effective optimized context retrieval (C1):} A reproducible and empirically validated pipeline for configuring optimized context retrieval systems is presented. This methodology, rigorously derived from a systematic analysis of key retrieval components, empowers researchers and practitioners to build efficient, high-performing systems affordably.
    \item \textbf{Empirically validated improved Pareto Frontier (C2):} A robust empirical validation of a significantly improved cost-accuracy Pareto frontier for open-source healthcare LLMs is provided. The optimized setup enables open models to achieve state-of-the-art accuracy at a reduced cost, demonstrably surpassing baseline open models and even challenging the efficiency of proprietary solutions.
    \item \textbf{OpenMedQA: A new benchmark for Open-Ended Medical QA (C3):} To address the limitations of multiple-choice medical QA, OpenMedQA is introduced, an extension of MedQA that evaluates open-ended medical question-answering capabilities. This benchmark enables direct comparison between MCQA and open-ended performance, highlighting the challenges of generating free-text medical responses. The findings reveal a significant accuracy gap, emphasizing the need for specific solutions in open-ended settings.
    \item \textbf{Open-source resources to empower healthcare AI development (C4):} To accelerate the development and adoption of healthcare AI solutions, a comprehensive suite of open-source resources are released to the community. This includes the \texttt{prompt\_engine} library, CoT/ToT/Thinking enhanced databases, and the OpenMedQA benchmark. These resources are aimed to empower the community to develop and deploy cost-optimized healthcare AI solutions and further explore the improved Pareto Frontier.
\end{itemize}

\section{Related Work}\label{sec:related_work}

Ensuring factuality in LLMs is a key focus in AI research, especially in healthcare, where reliability is critical~\cite{hager2024evaluation}. Initial efforts to enhance LLM accuracy have centered on In-Context Learning (ICL) techniques~\cite{brown2020languagemodelsfewshotlearners}. These methods, applied at inference time without reliance on external data sources, encompass sophisticated prompting strategies such as Chain of Thought (CoT)~\cite{wei2023chainofthoughtpromptingelicitsreasoning}, Tree of Thought (ToT)~\cite{yao2023treethoughtsdeliberateproblem}, Self-Consistency (SC)~\cite{wang2023selfconsistencyimproveschainthought}and Reasoning models~\cite{deepseekai2025deepseekr1incentivizingreasoningcapability}. While these techniques demonstrably improve response coherence and performance on intricate tasks, they often fall short of guaranteeing the stringent degree of factuality demanded in critical medical applications.

Retrieval Augmented Generation (RAG)~\cite{lewis2021retrievalaugmentedgenerationknowledgeintensivenlp} emerged as a more direct solution to the factuality problem, shifting the paradigm towards integrating external knowledge to bias LLM responses with reliable information. In healthcare, RAG-based approaches, often combined with prompting techniques, have shown considerable promise in improving accuracy on medical benchmarks~\cite{lievin2024can,pal2024gemini,Savage2024,Wang2024}. Medprompt~\cite{nori2023generalistfoundationmodelsoutcompete} stands out as a particularly sophisticated example, demonstrating state-of-the-art performance on medical MCQA tasks, integrating GPT-4 with an optimized RAG pipeline. Medprompt highlights the performance gains attainable through these strategies. However, the significant costs associated with deploying such a complex system, especially with expensive proprietary LLMs, hinder widespread accessibility.

In the last few years, a performance gap between large proprietary LLMs and their open-weight counterparts~\cite{lievin2024can} has persisted. However, recent progress in open-source LLM development~\cite{maharjan2024openmedlm,dubey2024llama3herdmodels,deepseekai2025deepseekr1incentivizingreasoningcapability}, combined with the potential of efficient RAG systems, suggests a promising pathway toward closing this gap while addressing the critical imperative of cost-effectiveness.

In this regard, Pareto frontiers provide a valuable framework for analyzing the inherent trade-off between model accuracy and computational cost~\cite{nori2024medprompto1explorationruntime}. Analyzing these frontiers requires quantifying the 'cost' axis, which reflects the resources needed for deployment. This cost can encompass factors like API calls, inference time, or computational effort. Estimating computational effort, often proxied by metrics like floating-point operations (FLOPs) or energy consumption, allows for a more direct comparison of the resource efficiency of different models and configurations. Calculating this efficiency dimension is crucial for assessing the practical deployability and scalability of LLM solutions in real-world healthcare settings, potentially impacting equitable access across diverse institutions.

Finally, the predominant reliance on Multiple-Choice Question Answering (MCQA) benchmarks for evaluation presents limitations~\cite{hager2024evaluation}. Real-world clinical practice necessitates handling open-ended questions requiring nuanced, generated responses.  While CoT prompting has been explored for open-ended QA tasks, research in this critical area remains comparatively limited and non-exhaustive~\cite{lievin2024can,Savage2024}.  

\section{Methodology}\label{sec:arch}

\begin{figure}[t]
  \centering
  \includegraphics[width=0.85\linewidth]{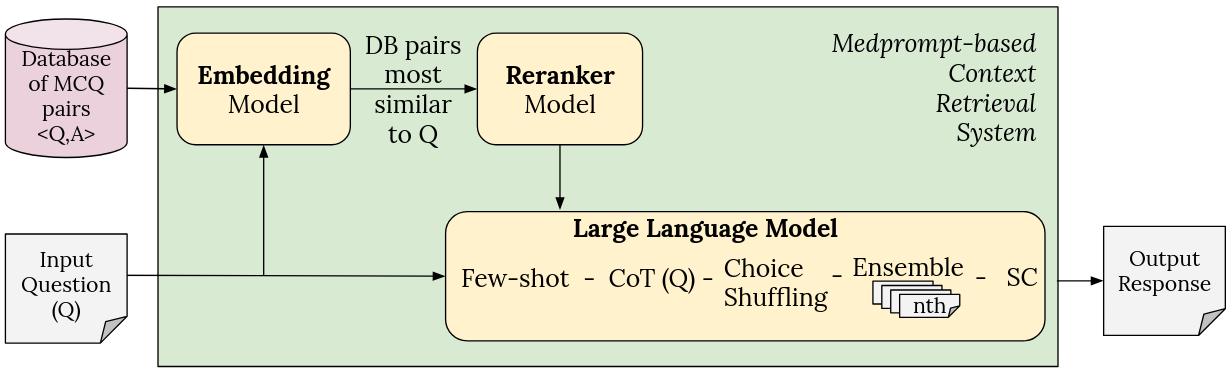}
  \caption{Components of the question-answering system based on context retrieval for LLMs.}\label{fig:arch}
\end{figure}

The retrieval architecture used in this work, and inspired by the Medprompt design, is illustrated in Figure~\ref{fig:arch}. This architecture effectively integrates Self-Consistency (SC) and Context Retrieval (CR) as key strategies to enhance the performance of LLMs, particularly medical MCQA benchmarks. CR is crucial for grounding the LLM in external knowledge, thereby improving factual accuracy, which is critical in healthcare applications. SC further refines the answer selection process by aggregating multiple reasoning paths, leading to more robust and reliable predictions.

The core components of this system are: (1) a database of examples, serving as the external knowledge source from which relevant medical information is retrieved; (2) an embedding model, responsible for encoding both queries and database text into a shared vector space to enable efficient semantic similarity search; and (3) a reranker, to further refine the retrieved context by re-ordering results based on more complex relevance criteria.  Beyond these retrieval-specific components, the system also incorporates choice shuffling to mitigate position bias in MCQA and optimizes ensemble size to achieve a balance between accuracy and computational cost. Each of these components significantly influences performance, and a systematic investigation was performed to optimize them in Section~\ref{sec:retrieval}, prioritizing a cost-effective and competitive configuration.

\subsection{Datasets and Models}\label{subsec:data}

Four different MCQA datasets from the healthcare domain were used to test the LLM-based question-answering system. \textbf{MedQA}~\cite{jin2021disease} consists of 1,273 USMLE-format questions. \textbf{MedMCQA}~\cite{pal2022medmcqalargescalemultisubject} includes 4,183 validation questions from Indian medical entrance exams, as the test set answers are private. \textbf{CareQA}~\cite{ariasduart2025automaticevaluationhealthcarellms} contains 5,621 questions from the Spanish Specialized Healthcare Training exam. \textbf{MMLU}~\cite{hendrycks2021measuringmassivemultitasklanguage} provides 1,089 medical-related questions from a multitask benchmark of 57 datasets, covering topics such as anatomy, clinical knowledge, and professional medicine. These datasets offer a diverse evaluation platform, covering various medical question styles and sources.

The main model used in our experiments is Llama3-Aloe-8B-Alpha~\cite{gururajan2024aloefamilyfinetunedopen}, a state-of-the-art open-source LLM specifically fine-tuned for the healthcare domain. It is used to evaluate the different components of the system, to find the optimal configuration (\S\ref{subsec:sccot_experiments} and \S\ref{subsec:medprompt_experiments}), which is then deployed to benchmark other models (\S\ref{subsec:sota_compare}). For this benchmark, the next generation of Aloe models (Aloe-Beta), Llama 3.1, and Qwen 2.5 model families are considered, as well as the DeepSeek-R1 reasoning model.

\section{Retrieval Experiments}\label{sec:retrieval}

This section presents an account of the experimental evaluation, designed to assess the impact of the Medprompt architecture  components (as outlined in Section~\ref{sec:arch}) on the performance of LLMs for medical MCQA. This investigation follows a structured approach: First, a study of the SC-CoT framework, examining the influence of its individual elements (\S\ref{subsec:sccot_experiments}). Second, the integration of external knowledge through context retrieval, analyzing the key components of the Medprompt architecture and their interplay (\S\ref{subsec:medprompt_experiments}). 
From these two component-wise analyses, an optimized CR configuration is derived and used to benchmark state-of-the-art LLMs to demonstrate the effectiveness and cost-efficiency of the proposed solution (\S\ref{subsec:sota_compare}).

\subsection{SC-CoT Experiments}\label{subsec:sccot_experiments}

Table \ref{tab:baselines} presents the baseline performance of Llama3-Aloe-8B-Alpha using zero-shot next token prediction, CoT, and SC-CoT. As hypothesized, SC-CoT consistently outperformed both zero-shot and standard CoT, demonstrating the benefit of aggregating multiple reasoning paths for improved answer selection. An analysis of the impact of choice shuffling, which mitigates LLM bias toward the first MCQ option~\cite{lievin2024can}, was also performed, showing consistent accuracy improvements across datasets (Table~\ref{tab:baselines}). Consequently, all subsequent experiments incorporated choice shuffling. 
 
\begin{table}[th]
\centering
\begin{tabular}{c|cccc}
   & \textbf{CareQA} & \textbf{MedMCQA} & \textbf{MedQA} &  \textbf{MMLU}\\
 \hline
\textbf{Zero-shot} & 67.57 & 58.91 & 62.45 & 72.76\\
\textbf{Zero-shot with CoT} & 65.11 & 55.10 & 64.26 & 72.93 \\
\hline
 \textbf{SC-CoT - 5 ensembles} & 67.64 &  56.78 &  64.81 & 73.68 \\
\textbf{+ Choice Shuffling} & +0.85 & +2.13 & +0.24 & +1.88 \\
\hline
\textbf{SC-CoT - 20 ensembles} & 68.89 & 56.78 &  64.10 &  73.79 \\
\textbf{+ Choice Shuffling} & +1.08 & +2.53 & +3.53 & +3.75 \\
\end{tabular}
\caption{Baseline accuracy of \textit{Llama3-Aloe-8B-Alpha} using 0-shot next-token prediction, CoT, and SC-CoT. The table also shows accuracy improvements when applying choice shuffling (CS) to SC-CoT.}
\label{tab:baselines}
\end{table}

\begin{figure}[t]
  \centering
  \includegraphics[width=   \linewidth]{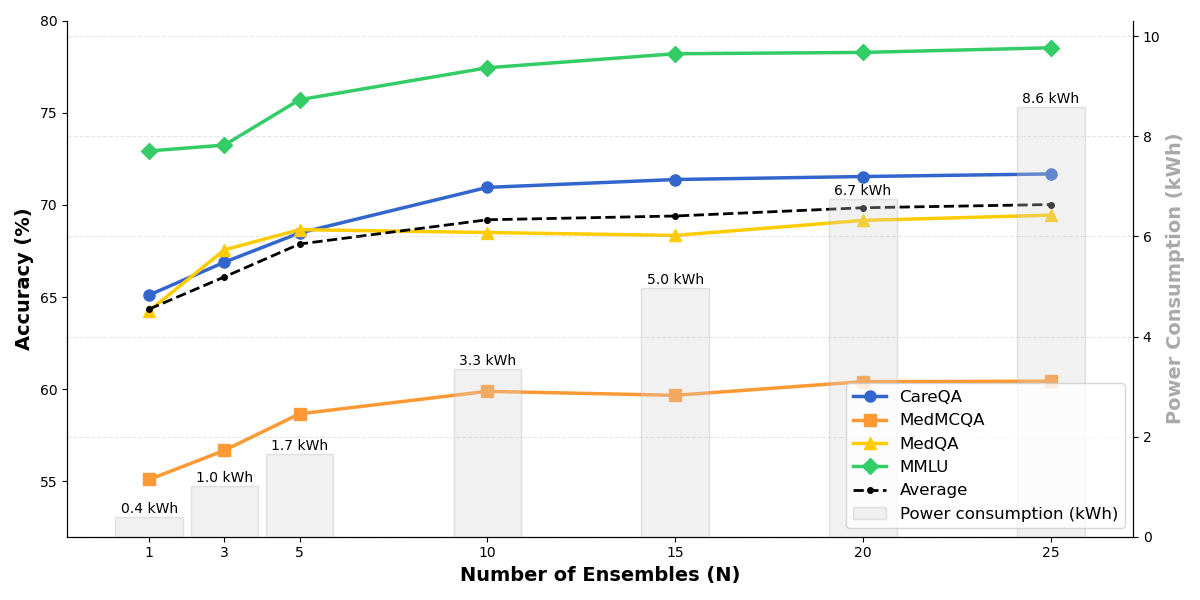}
  \caption{Accuracy vs. $CO_2$ emissions for increasing ensemble sizes in the SC-CoT setting. The solid lines represent accuracy trends for each dataset, while the dashed black line indicates the average accuracy. The shaded bars show power consumption in kWh, highlighting the trade-off between performance gains and environmental cost.}\label{fig:ens_trend}
\end{figure}

We then analyzed the impact of ensemble size ($N$) in SC-CoT. Here, $N$ represents the number of responses generated per question, with the final answer determined by majority voting among the selected options. As discussed in Section~\ref{sec:related_work}, evaluating cost-effectiveness requires analyzing the trade-off between accuracy gains and the associated computational resources. To quantify this for the ensemble size parameter, Figure~\ref{fig:ens_trend} plots accuracy against Estimated Energy Consumption (kWh). This metric serves as a proxy for the computational effort and associated operational cost, estimated based on model size, token throughput, and typical GPU power draw profiles during inference. Analyzing this relationship allows us to identify an optimal balance for $N$.

This trade-off can be seen in Figure~\ref{fig:ens_trend}, where performance as well as the footprint consistently have an upward trend with the first five ensembles yielding around 3.5\% accuracy gains. However, the additional ensembles provide minimal performance improvements at higher energy costs. Based on this, 5 ensembles were used for all the experiments, except in Table~\ref{tab:cr_vs_zs} where $N$ was set to 20 for maximal performance gains.

\subsection{Medprompt Experiments}\label{subsec:medprompt_experiments}

With an established robust baseline using optimized SC-CoT parameters, the impact of different choices for key components within the Medprompt architecture was investigated, aiming to identify a configuration that maximizes both accuracy and cost-effectiveness.

The first component of focus is the \textbf{embedding model}, which plays a critical role in retrieving relevant context from the external knowledge database. Four models with varying sizes and domain specializations were evaluated. Intriguingly, the results in Table~\ref{tab:embedding_res} reveal largely comparable performance across datasets, with no single embedding model exhibiting a clear and consistent advantage over the others across all benchmarks. It reveals that for medical MCQA, smaller, healthcare-specific embedding models like PubMedBERT achieve competitive retrieval quality. Considering both performance and computational efficiency, PubMedBERT was strategically selected as the default embedding model for subsequent experiments. However, for final state-of-the-art comparisons, the use of SFR-Mistral was also explored, given its slightly superior overall performance observed in this component analysis.

\begin{table}[th]
\centering
\small
\begin{tabular}{l|ccc|cccc}
\textbf{Model}  & \textbf{Dom.} & \textbf{Size} & \textbf{Param.} & \textbf{CareQA} & \textbf{MedMCQA} & \textbf{MedQA} &  \textbf{MMLU} \\
\hline
\textbf{PubMedBERT\cite{pubmedbertEmbedding}}   & Med. & 768  & 109M  & 68.65 & 59.55 & 69.60  & 75.55  \\
\textbf{MedCPT\cite{jin2023medcpt}}       & Med. & 768  & 109M & 68.81 & 59.29 & 67.16 &  75.44\\
\textbf{UAE-Large-V1\cite{li2023angle}}      & Gen. & 1024 & 335M & 68.08 & 59.53 & 69.05 &  76.70 \\
\textbf{SFR-Mistral\cite{SFRAIResearch2024}}      & Gen. & 4096 & 7B  & 68.61 & 60.60 & 70.15 &  73.33 \\
\end{tabular}
\caption{The table presents the characteristics of the embedding models and their performance in all datasets. $N$ and few-shot examples are set to 5, CS is activated, and the validation set of each dataset is used as the database of examples.}
\label{tab:embedding_res}
\end{table}

The second retrieval component evaluated was the \textbf{database}, which serves as the source from which relevant text excerpts are extracted and incorporated into the model prompt to provide additional context. To assess its impact, the original database method used in Medprompt was compared with the use of external databases. In Medprompt, the validation set of the datasets is used as the database by generating the answers at runtime. This approach has been replicated in \textit{prompt\_engine}. 

Additionally, this framework supports the use of a static, pre-created database. To this end, three distinct database types were constructed using different prompting strategies. The first two databases were generated using Llama-3.1-70B-instruct, employing CoT and ToT formats. For the CoT database, the model was prompted with the question, possible options, and the correct answer. It was instructed to analyze each option individually, provide detailed reasoning, and conclude by re-identifying the correct choice. For the ToT database, an adaption of the original ToT prompt is used to simulate three logical experts collaboratively answering the question. The third database was created using DeepSeek-R1. The reasoning model was prompted similarly to the CoT format, adding both the model's reasoning process and its final answer in the database. This approach aimed to simulate DeepSeek-R1's complex reasoning capabilities with other models. For generating these databases, we used the training sets from MedQA and MedMCQA. For MMLU and CareQA, which lack dedicated training sets, the MedMCQA database is used instead.

Table~\ref{tab:db} presents the results of these database ablation studies. The experiments on MedMCQA and MedQA directly evaluate the impact of database size and reasoning quality, while the experiments on CareQA and MMLU introduce a generalization challenge by utilizing a database derived from a different source (MedMCQA). The results in Table~\ref{tab:db} consistently demonstrate that larger databases, enriched with high-quality, reasoning-augmented data, generally lead to improved performance. Both CoT and ToT-augmented databases improved the accuracy in most datasets, surpassing the validation set database baseline. Interestingly, CoT-augmented databases exhibited slightly superior average performance compared to ToT.

However, the Thinking database, constructed using DeepSeek-R1, consistently yielded the most substantial performance gains across all datasets, achieving a noteworthy average improvement of 3.61\%. This finding strongly suggested that incorporating reasoning pathways distilled from a highly capable model like DeepSeek-R1 into the knowledge database is particularly effective in enhancing the performance of smaller LLMs, thus, the DeepSeek-R1 Thinking database was selected as the preferred knowledge source for the optimized context retrieval system.

Finally, the effectiveness of incorporating a \textbf{reranker} component was evaluated, specifically the MedCPT-Cross-Encoder~\cite{wang2023agree}, a specialized medical reranker. Contrary to the initial hypothesis, the inclusion of the MedCPT-Cross-Encoder reranker yielded inconsistent performance gains, and in some cases, even a slight performance degradation (\eg on CareQA and MMLU with CoT+Reranker in Table~\ref{tab:db}). Furthermore, the reranker introduced an additional computational overhead, increasing the overall inference latency. Given these inconsistent benefits and the added computational cost, the reranker was removed from the final optimized configuration, prioritizing efficiency and consistent performance.

\begin{table}[th]
\centering
\begin{tabular}{c|ccccc}
\textbf{Database}&\textbf{CareQA} & \textbf{MedMCQA} &\textbf{MedQA} &  \textbf{MMLU} & \textbf{Average} \\ \hline
\textbf{Validation set} & 68.65 &  59.55 & 69.60 &  75.55 & 68.34 \\
\textbf{Train+CoT} &  +0.78 & +7.15 &  -1.26  & +3.23 & +2.48\\
\textbf{Train+CoT+Reranker} & -0.18 & +0.10 & +1.02 & -1.35 & -0.10\\
\textbf{Train+ToT} & -1.83 & +4.40 & +0.63 & +1.88 & +1.27 \\
\textbf{Train+R1 think} & +0.43	& +7.88 & +2.36 &  +3.75 & +3.61 \\
\end{tabular}
\caption{Impact of database type and reasoning augmentation on medical MCQA accuracy. The table shows the accuracy change relative to the validation set baseline when extending the database using training splits augmented with CoT, ToT, and DeepSeek-R1 Thinking (R1 think) reasoning. Results also include experiments with reranking. Experiments use $N=5$ ensembles, 5 few-shot examples, choice shuffling, and PubMedBERT embedding model.}
\label{tab:db}
\end{table}

\subsection{State-of-the-art Comparison}\label{subsec:sota_compare}

The performance of the optimized CR system was benchmarked with various state-of-the-art LLMs, including models of different sizes and architectures. The setup follows the Medprompt scheme (Figure~\ref{fig:arch}), with key modifications: the exclusion of the reranker due to inconsistent performance gains and computational overhead, the use of SFR-Mistral for embeddings, DeepSeek-R1 Thinking-augmented training sets as the knowledge database, and 20 ensembles with choice shuffling to maximize performance.

\begin{table}[th]
\centering
\begin{tabular}{c|cccccc}
  \textbf{Model} & \textbf{CareQA} & \textbf{MedMCQA} & \textbf{MedQA} &  \textbf{MMLU} & \textbf{Average} \\
 \hline
 \textbf{Llama-3.1-8B} & 69.95 & 59.22 & 63.71 & 75.72 & 67.15 \\
 with CR  & +6.07 & +12.79 & +17.36 & +9.33 & +11.39  \\ \hline
 \textbf{Qwen2.5-7B} & 72.14 & 56.18	& 61.59 & 77.92 & 66.96\\ 
 with CR & +3.08 & +13.00 & +12.64 & +6.13 & +8.71 \\ \hline
 \textbf{Aloe-Beta-8B} & 70.77 &  59.57 & 64.65 & 76.50 & 67.87\\
 with CR & +5.37 & +12.72 & +16.26  & +7.60 & +10.49 \\\hline
 \textbf{Llama-3.1-70B} & 83.72 &  72.15 & 79.73  & 87.45 & 80.76\\ 
 with CR & +3.15 & +5.69 & +9.66 & +3.84 & +5.54 \\ \hline
  \textbf{Qwen2.5-72B} & 85.45 &  69.26 & 77.85 & 88.81 & 80.34 \\
 with CR & +1.08 & +7.55 & +7.46 & +2.75 & +4.71 \\ \hline
 \textbf{Aloe-Beta-70B} &  83.19 &  72.15 & 79.73 & 88.44  & 80.88 \\ 
 with CR & +4.38 & +5.28 & +9.11 & +3.01 & +5.45 \\ \hline
 \textbf{DeepSeek-R1} & 88.33 & 73.34 & 82.48  & 91.27 & 83.86 \\ 
 with CR & +4.18 & +8.94 & +11.94 & +3.61 & +7.17 \\ 
 \hline
\multicolumn{6}{c}{\textit{Private models}} \\
\textbf{GPT-4 + Medprompt*} & -  & 79.10 & 90.20 & 94.2 & -  \\ 
\textbf{MedPalm-2  + ER*} & - & 72.30 & 85.40 & 89.40 & - \\ 
\textbf{O1 + TPE*} & - &  83.90 & 96.00 & 95.28 & - \\
\end{tabular}\caption{Benchmarking state-of-the-art LLMs with and without optimized context retrieval (CR) on medical MCQA benchmarks. \\
\textbf{*} Results reported by others~\cite{nori2023generalistfoundationmodelsoutcompete,singhal2023expertlevelmedicalquestionanswering,nori2024medprompto1explorationruntime}. ER: Ensemble Refinement (Google's custom prompt technique). TPE: Tailored Prompt Ensemble (custom OpenAI ensemble technique).}
\label{tab:cr_vs_zs}
\end{table}

Table~\ref{tab:cr_vs_zs} presents the comparative performance of various LLMs, evaluated both in their zero-shot configuration and when augmented with the optimized CR system. The results unequivocally demonstrate the effectiveness of the optimized CR approach in enhancing the performance of LLMs on medical MCQA tasks. Across all datasets and models assessed, the addition of context retrieval consistently yielded statistically significant accuracy improvement. Notably, the magnitude of the gains exhibited an inverse correlation with the base performance of the model. Smaller models exhibited the most substantial relative improvements, with average accuracy gains exceeding 10\%. This suggests that context retrieval is particularly effective in compensating for the limited inherent knowledge of smaller LLMs, enabling them to achieve performance levels that approach or even rival those of much larger models. Even for high-performing models such as DeepSeek-R1, context retrieval provided a significant boost of over 7\% average accuracy, further pushing the state-of-the-art.

Most importantly, the results demonstrate that optimized context retrieval provides a powerful pathway to achieve \textbf{cost-effective high performance} in healthcare AI. This cost reduction is visually and quantitatively substantiated by the Pareto frontier analysis presented in Figure~\ref{fig:pareto}, which demonstrably illustrates the shift toward a more efficient accuracy-cost trade-off enabled by the approach. By leveraging open-source models like Aloe-Beta-70B and DeepSeek-R1 in conjunction with optimized CR, state-of-the-art performance is achieved while drastically reducing operational costs. The cost was estimated by multiplying the total tokens with costs from the most affordable provider as of date \footnote{Artifiical Analysis: \url{https://artificialanalysis.ai/}}, while the values for the closed models were taken from~\cite{nori2024medprompto1explorationruntime}. This finding has significant practical implications for enhancing accessibility access to high-quality healthcare AI, particularly in resource-constrained environments, where cost-effectiveness is a paramount consideration. 

\section{OpenMedQA}\label{subsec:beyond_qa}

While MCQA benchmarks are instrumental in evaluating LLMs for medical applications, they fall short of capturing the complexities of real-world clinical reasoning. In practice, healthcare professionals must generate comprehensive responses without relying on predefined answer choices. This highlights a critical need for evaluating Open-Ended Question-Answering (OE-QA) capabilities in medical AI systems, as these systems must handle the inherent variability and nuance found in clinical practice.

Evaluating OE medical questions presents unique challenges compared to MCQA. Unlike multiple-choice formats, OE responses require the model to produce detailed explanations, consider context, and respond to complex queries without guidance from predetermined options. This makes the assessment of OE-QA more demanding, as it must account for diverse yet clinically valid responses. In addition, the lack of standardized evaluation metrics for OE answers further complicates the process, underscoring the importance of developing robust frameworks for their assessment.

To bridge this evaluation gap, \textbf{OpenMedQA} is presented, a novel benchmark designed for open-ended medical question-answering. OpenMedQA builds upon the MedQA dataset, enabling a direct comparison between MCQA and OE-QA formats. The MedQA questions were rephrased using the DeepSeek-R1 model to maintain their original medical intent, and the resulting dataset is made publicly available to support research. For the rephrased questions, the answer corresponds solely to the correct option from the original MCQA format, ensuring that the dataset remains grounded in verified medical knowledge. Out of the original 1,273 questions, 1,272 rephrased open-ended questions were obtained, removing one question that originally required an image to answer it (index 454 of the original test set).

Model performance on both the MC and OE versions were compared using several LLMs. For MCQA, standard evaluation methods, such as computing the log-likelihood of each option or parsing the model’s output to determine the selected choice (A, B, C, or D) were used. However, these approaches are not directly applicable to OE-QA, where models generate free-form text responses rather than selecting from predefined options.

To address this challenge, an LLM-as-a-judge approach was employed for automated evaluation. Specifically, DeepSeek-R1 was leveraged, which has demonstrated strong reasoning capabilities, to assess the quality of model-generated responses. The evaluation process involves comparing each response against ground truth references, considering both factual correctness and clinical relevance. By using a dedicated state-of-the-art LLM as an evaluator, a reliable framework for assessing open-ended medical QA performance was established in an objective manner.

\begin{table}[h]
\centering
\begin{tabular}{l|cc|c}
\textbf{Model} & \textbf{ MedQA } & \textbf{ OpenMedQA } & \textbf{Performance Drop} \\
\hline
Llama-3.1-8B-Instruct     & 63.71 & 33.88 & -29.82\\ \hline
Qwen2.5-7B-Instruct       & 61.59 & 38.76 & -22.83 \\ \hline
Llama3.1-Aloe-Beta-8B     & 64.65 & 52.91 & -11.74 \\ \hline
Llama-3.1-70B-Instruct    & 79.73 & 60.46 & -19.28 \\ \hline
Qwen2.5-72B-Chat          & 77.85 & 61.24 & -16.61 \\ \hline
Llama3.1-Aloe-Beta-70B    & 79.73 & 65.02 & -14.72 \\ \hline
DeepSeek-R1               & 82.48 & 75.86 & -6.62 \\  
\end{tabular}
\caption{Comparison of multiple-choice (MedQA) and open-ended (OpenMedQA) question-answering accuracy (in \%) across various models, highlighting the performance gap between both formats.}
\label{tab:open_baselines}
\end{table}

Table \ref{tab:open_baselines} presents the accuracy results of various models on MC and OE settings, along with the corresponding performance drop when shifting to the OE format. The results indicate a consistent decline in accuracy across all models when moving from MC to OE evaluation, highlighting the increased difficulty of generating free-text responses compared to selecting from predefined options. Smaller models exhibit the largest performance drops, suggesting that limited model capacity affects their ability to generate high-quality, well-structured responses in an open-ended setting. Conversely, models that have undergone instruction tuning, such as Aloe-Beta variants, show a relatively smaller drop, indicating that specific domain fine-tuning may improve medical OE reasoning capabilities.

Among the larger models, DeepSeek-R1 performs the best with a modest performance drop, suggesting superior generalization and reasoning abilities in open-ended tasks. However, this result might be influenced by potential bias, as DeepSeek-R1 also serves as the judge. Meanwhile, models such as Llama-3.1-70B and Qwen2.5-72B experience noticeable declines, demonstrating that even state-of-the-art models struggle with generating free-text medical answers. These findings emphasize the need for improved evaluation methods and training strategies to bridge the gap between structured multiple-choice formats and the more complex OE-QA tasks required in real-world clinical applications. Furthermore, the results suggest that while increasing model size generally leads to better performance, scaling alone is insufficient to fully close the gap between MCQA and OE-QA.

\section{Conclusions}

This work underscores the significant potential of augmenting LLMs with CR systems to enhance their accuracy and reliability in the healthcare domain. Our exploration of SC-CoT components revealed substantial gains through choice shuffling and an optimal number of ensembles, striking a balance between performance and computational cost. Further investigation into the Medprompt architecture highlighted the effectiveness of small, healthcare-specific embedding models and the value of enriching the prompt with high-quality knowledge can bridge the performance gap between smaller and larger models, enabling cost-effective and reliable solutions for healthcare applications.  

By incorporating reasoning pathways from superior models like DeepSeek-R1, smaller LLMs were able to emulate the advanced problem-solving processes, achieving accuracy levels that rival or surpass proprietary models. For instance, DeepSeek-R1 with CR achieves an accuracy of 92.51\% on CareQA, representing the highest reported performance to date. This approach not only improves performance but also redefines the Pareto frontier on the MedQA benchmark, enabling open-source models to operate at a fraction of the cost while maintaining state-of-the-art accuracy.

Extending this framework to OE-QA, OpenMedQA was introduced, a novel benchmark derived from MedQA, to rigorously evaluate open-ended medical question answering. The results highlight a consistent performance drop when transitioning from MCQA to OE-QA, demonstrating the increased difficulty of generating free-text medical responses. Even state-of-the-art models experience a notable decline in accuracy, underscoring the challenges of OE reasoning in medical AI. Among all evaluated models, DeepSeek-R1 exhibited the smallest performance drop, suggesting its superior ability to generalize in free-text scenarios. These findings reinforce the need for more effective strategies in training and evaluating medical AI systems, including improved retrieval-augmented generation techniques and reasoning-guided prompting.

Future work should focus on refining CR strategies to further enhance LLM reasoning across both MCQA and OE-QA tasks. While the results demonstrate strong performance in multiple-choice settings, the significant drop in accuracy for open-ended responses highlights the need for better generation techniques. Future research could explore adaptive ensemble methods tailored for OE-QA, leveraging reasoning chains optimized for free-text generation. Additionally, integrating domain-specific retrieval mechanisms and reinforcement learning could help bridge the performance gap, ensuring more reliable and interpretable medical AI models.

\subsubsection{Acknowledgements}
This work is supported by Jordi Bayarri fellowship within the “Generación D” initiative, \href{https://www.red.es/es}{Red.es}, Ministerio para la Transformación Digital y de la Función Pública, for talent attraction (C005/24-ED CV1). Funded by the European Union NextGenerationEU funds, through PRTR. We also acknowledge the computational resources provided by the FinisTerrae III, Leonardo, and MareNostrum 5 supercomputers. We are particularly grateful to the Operations department at BSC for their technical support.
%
%
%
\bibliographystyle{splncs04}
\bibliography{references}

\end{document}